%% file: icmla_2018.tex
\documentclass[conference]{IEEEtran}
\IEEEoverridecommandlockouts
\usepackage{amsmath,amssymb,amsfonts}
\usepackage{endnotes}
\usepackage{algorithmic}
\usepackage{graphicx}
\usepackage{textcomp}
\usepackage{xcolor}
\usepackage{color,soul}
\usepackage{booktabs, tabularx, caption}
\usepackage{hyperref}
\usepackage{subfiles}
\usepackage{fancyhdr}
\usepackage{eso-pic}

\def\BibTeX{{\rm B\kern-.05em{\sc i\kern-.025em b}\kern-.08em
    T\kern-.1667em\lower.7ex\hbox{E}\kern-.125emX}}
\begin{document}

\AddToShipoutPictureBG*{%
  \AtPageLowerLeft{%
    \setlength\unitlength{1in}%
    \hspace*{\dimexpr0.5\paperwidth\relax}
    \makebox(0,0.75)[c]{17th IEEE International Conference on Machine Learning and Applications (IEEE ICMLA 2018), Orlando, Florida, USA}%
}}

\renewcommand{\footnoterule}{%
\kern -3pt
\hrule width 2in
\kern 2.6pt
}

\title{Automated Vulnerability Detection in Source Code Using Deep Representation Learning}


\newcommand*{\affmark}[1][*]{\textsuperscript{#1}}

\author{
  {\rm Rebecca L. Russell\affmark[1]\textsuperscript{*}\thanks{\textsuperscript{*}Correspondence to: rrussell@draper.com}, Louis Kim\affmark[1], Lei H. Hamilton\affmark[1], Tomo Lazovich\affmark[1]\textsuperscript{\dag}\thanks{\textsuperscript{\dag}Tomo Lazovich now works at Lightmatter},} \\
  {\rm Jacob A. Harer\affmark[1,2], Onur Ozdemir\affmark[1], Paul M. Ellingwood\affmark[1], Marc W. McConley\affmark[1]} \\ \\
  \affmark[1] \textit{Draper}\\
  \affmark[2] \textit{Boston University}
}

\maketitle

\begin{abstract}
\subfile{sections/abstract}
\end{abstract}


\section{Introduction}
\subfile{sections/introduction}

\section{Related Work}
\subfile{sections/related-work}

\section{Data}
\subfile{sections/data}

\section{Methods}

\subfile{sections/methods}

\section{Results}

\subfile{sections/results}

\section{Conclusions}
\subfile{sections/conclusions}

\section*{Acknowledgment}
The authors thank Hugh J. Enxing and Thomas Jost for their efforts creating the data ingestion pipeline. This project was sponsored by the Air Force Research Laboratory (AFRL) as part of the DARPA MUSE program.

\bibliographystyle{ieeetr}
\bibliography{references} 
\end{document}

%% file: sections/abstract.tex
Increasing numbers of software vulnerabilities are discovered every year whether they are reported publicly or discovered internally in proprietary code. These vulnerabilities can pose serious risk of exploit and result in system compromise, information leaks, or denial of service. We leveraged the wealth of C and C++ open-source code available to develop a large-scale function-level vulnerability detection system using machine learning. To supplement existing labeled vulnerability datasets, we compiled a vast dataset of millions of open-source functions and labeled it with carefully-selected findings from three different static analyzers that indicate potential exploits. The labeled dataset is available at: \url{https://osf.io/d45bw/}. Using these datasets, we developed a fast and scalable vulnerability detection tool based on deep feature representation learning that directly interprets lexed source code. We evaluated our tool on code from both real software packages and the NIST SATE IV benchmark dataset. Our results demonstrate that deep feature representation learning on source code is a promising approach for automated software vulnerability detection.

%% file: sections/introduction.tex
Hidden flaws in software can result in security vulnerabilities that potentially allow attackers to compromise systems and applications. Thousands of such vulnerabilities are reported publicly to the Common Vulnerabilities and Exposures database~\cite{cwe} each year and many more are discovered internally in proprietary code and patched. Recent high-profile exploits have shown that these security holes can have disastrous effects, both financially and societally~\cite{equifax}. These vulnerabilities are often caused by subtle errors made by programmers and can propagate quickly due to the prevalence of open-source software and code reuse.

While there are existing tools for program analysis, these tools typically only detect a limited subset of possible errors based on pre-defined rules. With the recent widespread availability of open-source repositories, it has become possible to use data-driven techniques to discover vulnerability patterns. We present machine learning (ML) techniques for the automated detection of vulnerabilities in C/C++\footnote{While our work focuses on C/C++, the techniques are applicable to any programming language.} source code learned from real world code examples.

%% file: sections/related-work.tex
There currently exist a wide variety of analysis tools that attempt to uncover common vulnerabilities in software. Static analyzers, such as Clang \cite{Xu_Clang}, do so without needing to execute programs. Dynamic analyzers repeatedly execute programs with many test inputs on real or virtual processors to identify weaknesses. Both static and dynamic analyzers are rule-based tools and thus limited to their hand-engineered rules and not able to guarantee full test coverage of codebases. Symbolic execution \cite{King} replaces input data with symbolic values and analyzes their use over the control flow graph of a program. While it can probe all feasible program paths, symbolic execution is expensive and does not scale well to large programs.

Beyond these traditional tools, there has been significant recent work on the usage of machine learning for program analysis. The availability of large amounts of open-source code opens the opportunity to learn the patterns of software vulnerabilities directly from mined data. For a comprehensive review of learning from ``Big Code'', including that not directly related to our work, see Allamanis et al. \cite{Allamanis2017}.

In the area of vulnerability detection, Hovsepyan et al. \cite{Hovsepyan} used a support vector machine (SVM) on a bag-of-words (BOW) representation of a simple tokenization of Java source code to predict static analyzer labels. However, their work was limited to training and evaluating on a single software repository.
Pang et al. \cite{Pang} expanded on this work by including n-grams in the feature vectors used with the SVM classifier.
Mou et al. \cite{Mou} explored the potential of deep learning for program analysis by embedding the nodes of the abstract syntax tree representations of source code and training a tree-based convolutional neural network for simple supervised classification problems.
Li et al. \cite{VulDeePecker} used a recurrent neural network (RNN) trained on code snippets related to library/API function calls to detect two types of vulnerabilities related to the improper usage of those calls.
Harer et al. \cite{harer2018} trained an RNN to detect vulnerabilities in the lexed representations of functions in a synthetic codebase, as part of a generative adversarial approach to code repair.

To our knowledge, no work has been done on using deep learning to learn features directly from source code in a large natural codebase to detect a variety of vulnerabilities. The limited datasets (in both size and variety) used by most of the previous works limit the usefulness of the results and prevent them from taking full advantage of the power of deep learning.

%% file: sections/data.tex
Given the complexity and variety of programs, a large number of training examples are required to train machine learning models that can effectively learn the patterns of security vulnerabilities directly from code. We chose to analyze software packages at the function-level because it is the lowest level of granularity capturing the overall flow of a subroutine. We compiled a vast dataset of millions of function-level examples of C and C++ code from the SATE IV Juliet Test Suite~\cite{sate4juliet}, Debian Linux distribution~\cite{debian}, and public Git repositories on GitHub~\cite{github}. Table~\ref{table:data} shows the data summary of the number of functions we collected and used from each source in our dataset of over 12 million functions.

\begin{table}[!htbp]
\begin{center}
\begin{tabular}{lcccc}
\toprule
& SATE IV & GitHub & Debian \\
\midrule
Total &  121,353  & 9,706,269 & 3,046,758 \\
Passing curation &   11,896  & 782,493 & 491,873 \\
~~ `Not vulnerable' & 6,503 (55{\%})  & 730,160 (93{\%}) & 461,795 (94{\%}) \\
~~ `Vulnerable' &  5,393 (45{\%}) & 52,333 (7{\%}) & 30,078 (6{\%}) \\
\bottomrule
\end{tabular}
\caption{Total number of functions obtained from each data source, the number of valid functions remaining after removing duplicates and applying cuts, and the number of functions without and with detected vulnerabilities.}
\label{table:data}
\end{center}
\end{table}

The SATE IV Juliet Test Suite contains synthetic code examples with vulnerabilities from 118 different Common Weakness Enumeration (CWE)~\cite{cwe} classes and was originally designed to explore the performance of static and dynamic analyzers. While the SATE IV dataset provides labeled examples of many types of vulnerabilities, it is made up of synthetic code snippets that do not sufficiently cover the space of natural code to provide an appropriate training set alone. To provide a vast dataset of natural code to augment the SATE IV data, we mined large numbers of functions from Debian packages and public Git repositories. The Debian package releases provide a selection of very well-managed and curated code which is in use on many systems today. The GitHub dataset provides a larger quantity and wider variety of (often lower-quality) code. Since the open-source functions from Debian and GitHub are not labeled, we used a suite of static analysis tools to generate the labels. Details of the label generation are explained in Subsection~\ref{ssec:labels}.

\begin{table*}[!htbp]
\begin{center}
\begin{tabular}{clc}

\toprule
CWE ID& CWE Description & Frequency {\%} \\
\midrule
120/121/122 &  Buffer Overflow  & 38.2{\%}  \\
119 & Improper Restriction of Operations within the Bounds of a Memory Buffer & 18.9{\%}     \\
476 & NULL Pointer Dereference & 9.5{\%}      \\
469 & Use of Pointer Subtraction to Determine Size & 2.0{\%}  \\
20, 457, 805 etc. & Improper Input Validation, Use of Uninitialized Variable, Buffer Access with Incorrect Length Value, etc. & 31.4{\%} \\
\bottomrule
\end{tabular}
\caption{CWE statistics of vulnerabilities detected in our C/C++ dataset.}
\label{table:bug_types}
\end{center}
\end{table*}

\subsection{Source lexing}
To generate useful features from the raw source code of each function, we created a custom C/C++ lexer designed to capture the relevant meaning of critical tokens while keeping the representation generic and minimizing the total token vocabulary size. Making our lexed representation of code from different software repositories as standardized as possible empowers transfer learning across the full dataset. Standard lexers, designed for actually compiling code, capture far too much detail that can lead to overfitting in ML approaches.

Our lexer was able to reduce C/C++ code to representations using a total vocabulary size of only 156 tokens. All base C/C++ keywords, operators, and separators are included in the vocabulary. Code that does not affect compilation, such as comments, is stripped out. String, character, and float literals are lexed to type-specific placeholder tokens, as are all identifiers. Integer literals are tokenized digit-by-digit, as these values are frequently relevant to vulnerabilities. Types and function calls from common libraries that are likely to have relevance to vulnerabilities are mapped to generic versions. For example, \texttt{u32}, \texttt{uint32\_t}, \texttt{UINT32}, \texttt{uint32}, and \texttt{DWORD} are all lexed as the same generic token representing 32-bit unsigned data types. Learned embeddings of these individual tokens would likely distinguish them based on the kind of code they are commonly used in, so care was taken to build in the desired invariance.

\subsection{Data curation}
One very important step of our data preparation was the removal of potential duplicate functions. Open-source repositories often have functions duplicated across different packages. Such duplication can artificially inflate performance metrics and conceal overfitting, as training data can leak into test sets. Likewise, there are many functions that are \emph{near} duplicates, containing trivial changes in source code that do not significantly affect the execution of the function. These near duplicates are challenging to remove, as they can often appear in very different code repositories and can look quite different at the raw source level.

To protect against these issues, we performed an extremely strict duplicate removal process. We removed any function with a duplicated lexed representation of its source code {\textit{or}} a duplicated compile-level feature vector. This compile-level feature vector was created by extracting the control flow graph of the function as well as the operations happening in each basic block (opcode vector, or op-vec) and the definition and use of variables (use-def matrix)\footnote{Our compile-level feature extraction framework incorporated modified variants of strace and buildbot as well as a custom Clang plugin -- we omit the details to focus on the ML aspects of our work.}. Two functions with identical instruction-level behaviors or functionality are likely to have both similar lexed representations and highly correlated vulnerability status.

The ``Passing curation'' row of Table~\ref{table:data} reflects the number of functions remaining after the duplicate removal process, about 10.8{\%} of the total number of functions pulled. Although our strict duplicate removal process filters out a significant amount of data, this approach provides the {\textit{most conservative}} performance results, closely estimating how well our tool will perform against code it has never seen before.

\subsection{Labels} \label{ssec:labels}
Labeling code vulnerability at the function level was a significant challenge. The bulk of our dataset was made up of mined open-source code without known ground truth. In order to generate labels, we pursued three approaches: static analysis, dynamic analysis, and commit-message/bug-report tagging. 

While dynamic analysis is capable of exposing subtle flaws by executing functions with a wide range of possible inputs, it is extremely resource intensive. Performing a dynamic analysis on the roughly 400 functions in a single module of the LibTIFF 3.8.2 package from the ManyBugs dataset~\cite{ManyBugs} took nearly a day of effort. Therefore, this approach was not realistic for our extremely large dataset.

Commit-message based labeling turned out to be very challenging, providing low-quality labels. In our tests, both humans and ML algorithms were poor at using commit messages to predict corresponding Travis CI~\cite{travis} build failures or fixes. Motivated by recent work by Zhou et al.~\cite{Zhou}, we also tried a simple keyword search looking for commit words like ``buggy'', ``broken'', ``error'', ``fixed'', etc. to label before-and-after pairs of functions, which yielded better results in terms of relevancy. However, this approach greatly reduced the number of candidate functions that we could label and still required significant manual inspection, making it inappropriate for our vast dataset.

As a result, we decided to use three open-source static analyzers, Clang, Cppcheck~\cite{cppcheck}, and Flawfinder~\cite{flawfinder}, to generate labels. Each static analyzer varies in its scope of search and detection. For example, Clang's scope is very broad but also picks up on syntax, programming style, and other findings which are not likely to result in a vulnerability. Flawfinder's scope is geared towards CWEs and does not focus on other aspects such as style. Therefore, we incorporated multiple static analyzers and pruned their outputs to exclude findings that are not typically associated with security vulnerabilities in an effort to create robust labels.

We had a team of security researchers map each static analyzer's finding categories to the corresponding CWEs and identify which CWEs would likely result in potential security vulnerabilities. This process allowed us to generate binary labels of ``vulnerable'' and ``not vulnerable'', depending on the CWE. For example, Clang's ``Out-of-bound array access'' finding was mapped to ``CWE-805: Buffer Access with Incorrect Length Value'', an exploitable vulnerability that can lead to program crashes, so functions with this finding were labeled ``vulnerable''. On the other hand, Cppcheck's ``Unused struct member'' finding was mapped to ``CWE-563: Assignment to Variable without Use'', a poor code practice unlikely to cause a security vulnerability, so corresponding functions were labeled ``not vulnerable'' even though static analyzers flagged them. Of the 390 total types of findings from the static analyzers, 149 were determined to result in a potential security vulnerability. Roughly 6.8\% of our curated, mined C/C++ functions triggered a vulnerability-related finding. Table~\ref{table:bug_types} shows the statistics of frequent CWEs in these ``vulnerable'' functions. All open-source function source codes from Debian and GitHub with corresponding CWE labels are available here: \url{https://osf.io/d45bw/}.

%% file: sections/methods.tex
\begin{figure*}[!htbp]
\includegraphics[width=\textwidth]{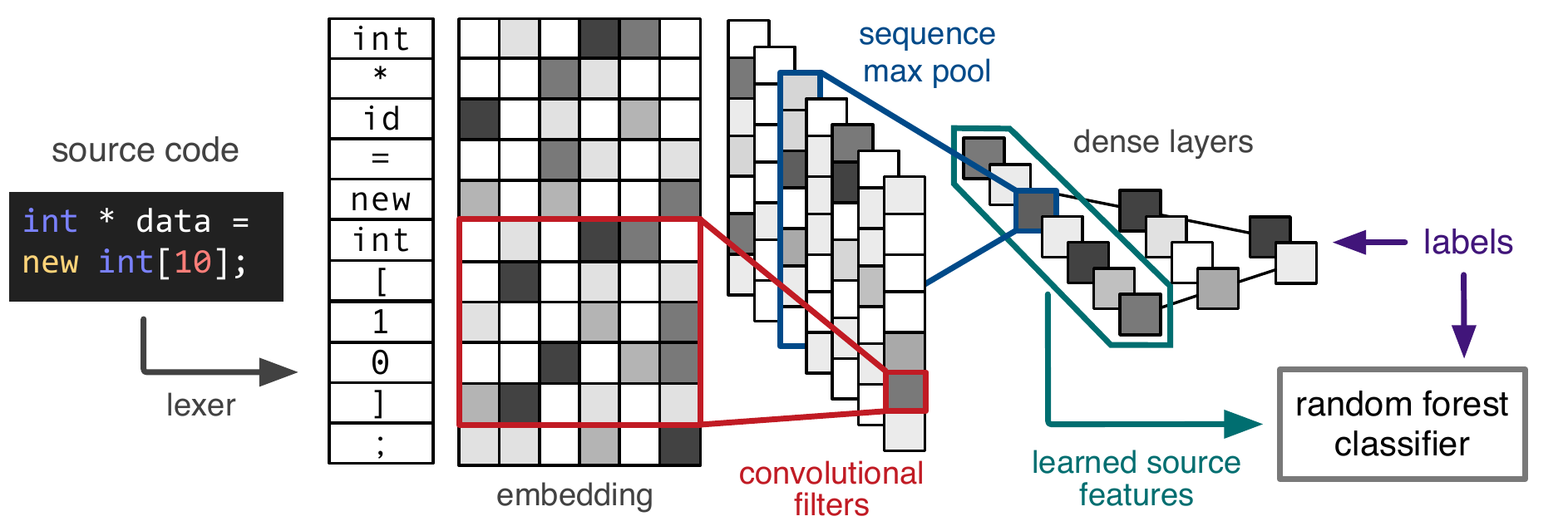}
\caption{Illustration of our convolutional neural representation-learning approach to source code classification. Input source code is lexed into a token sequence of variable length $\ell$, embedded into a $\ell\times k$ representation, filtered by $n$ convolutions of size $m\times k$, and maxpooled along the sequence length to a feature vector of fixed size $n$. The embedding and convolutional filters are learned by weighted cross entropy loss from fully-connected classification layers. The learned $n$-dimensional feature vector is used as input to a random forest classifier, which improves performance compared to the neural network classifier alone.}
\label{fig:source_method}
\end{figure*}

Our primary machine learning approach to vulnerability detection, depicted in Figure \ref{fig:source_method}, combines the neural feature representations of lexed function source code with a powerful ensemble classifier, random forest (RF).

\subsection{Neural network classification and representation learning}
Since source code shares some commonalities with writing and work done for programming languages is more limited, we build off approaches developed for natural language processing (NLP) \cite{kim2014}.
We leverage feature-extraction approaches similar to those used for sentence sentiment classification with convolutional neural networks (CNNs) and recurrent neural networks (RNNs) for function-level source vulnerability classification. 
\subsubsection{Embedding}
The tokens making up the lexed functions are first embedded into a fixed $k$-dimensional representation (limited to range $[-1, 1]$) that is learned during classification training via backpropagation to a linear transformation of a one-hot embedding. Several unsupervised word2vec approaches \cite{mikolov2013} trained on a much larger unlabeled dataset were explored for seeding this embedding, but these yielded minimal improvement in classification performance over randomly-initialized learned embeddings. A fixed one-hot embedding was also tried, but gave diminished results. As our vocabulary size is much smaller than those of natural languages, we were able to use a much smaller embedding than is typical in NLP applications. Our experiments found that $k=13$ performed the best for supervised embedding sizes, balancing the expressiveness of the embedding against overfitting. We found that adding a small amount of random Gaussian noise $\mathcal{N}\left(\mu=0,\sigma^2=0.01\right)$ to each embedded representation substantially improved resistance to overfitting and was much more effective than other regularization techniques such as weight decay.

\subsubsection{Feature extraction}
We explored both CNNs and RNNs for feature extraction from the embedded source representations.
\textbf{Convolutional feature extraction}: We use $n$ convolutional filters with shape $m\times k$, so each filter spans the full space of the token embedding. The filter size $m$ determines the number of sequential tokens that are considered together and we found that a fairly large filter size of $m=9$ worked best. A total number of $n=512$ filters, paired with batch normalization followed by ReLU, was most effective.
\textbf{Recurrent feature extraction}: We also explored using recurrent neural networks for feature extraction to allow longer token-dependencies to be captured. The embedded representation is fed to a multi-layer RNN and the output at each step in the length $\ell$ sequence is concatenated. We used two-layer Gated Recurrent Unit RNNs with hidden state size $n'=256$, though Long Short Term Memory RNNs performed equally well.
\subsubsection{Pooling}
As the length of C/C++ functions found in the wild can vary dramatically, both the convolutional and recurrent features are maxpooled along the sequence length $\ell$ in order to generate a fixed-size ($n$ or $n'$, respectively) representation. In this architecture, the feature extraction layers should learn to identify different signals of vulnerability and thus the presence of any of these along the sequence is important.
\subsubsection{Dense layers}
The feature extraction layers are followed by a fully-connected classifier. 50\% dropout on the maxpooled feature representation connections to the first hidden layer was used when training. We found that using two hidden layers of 64 and 16 before the final softmax output layer gave the best classification performance.
\subsubsection{Training}
For data batching convenience, we trained only on functions with token length $10 \leq \ell \leq 500$, padded to the maximum length of $500$. Both the convolutional and recurrent networks were trained with batch size 128, Adam optimization (with learning rates $5\times10^{-4}$ and $1\times10^{-4}$, respectively), and with a cross entropy loss. Since the dataset was strongly unbalanced, vulnerable functions were weighted more heavily in the loss function. This weight is one of the many hyper-parameters we tuned to get the best performance. We used a 80:10:10 split of our SATE IV, Debian, and GitHub combined dataset to train, validate, and test our models. We tuned and selected models based on the highest validation Matthews Correlation Coefficient (MCC).

\subsection{Ensemble learning on neural representations}
While the neural network approaches automatically build their own features, their classification performance on our full dataset was suboptimal. We found that using the neural features (outputs from the sequence-maxpooled convolution layer in the CNN and sequence-maxpooled output states in the RNN) as inputs to a powerful ensemble classifier such as random forest or extremely randomized trees yielded the best results on our full dataset. Having the features and classifier optimized separately seemed to help resist overfitting. This approach also makes it more convenient to quickly retrain a classifier on new sets of features or combinations of features.

%% file: sections/results.tex
\begin{figure*}[!tbp]
\centering
\begin{minipage}[t]{0.48\textwidth}
\includegraphics[width=\textwidth]{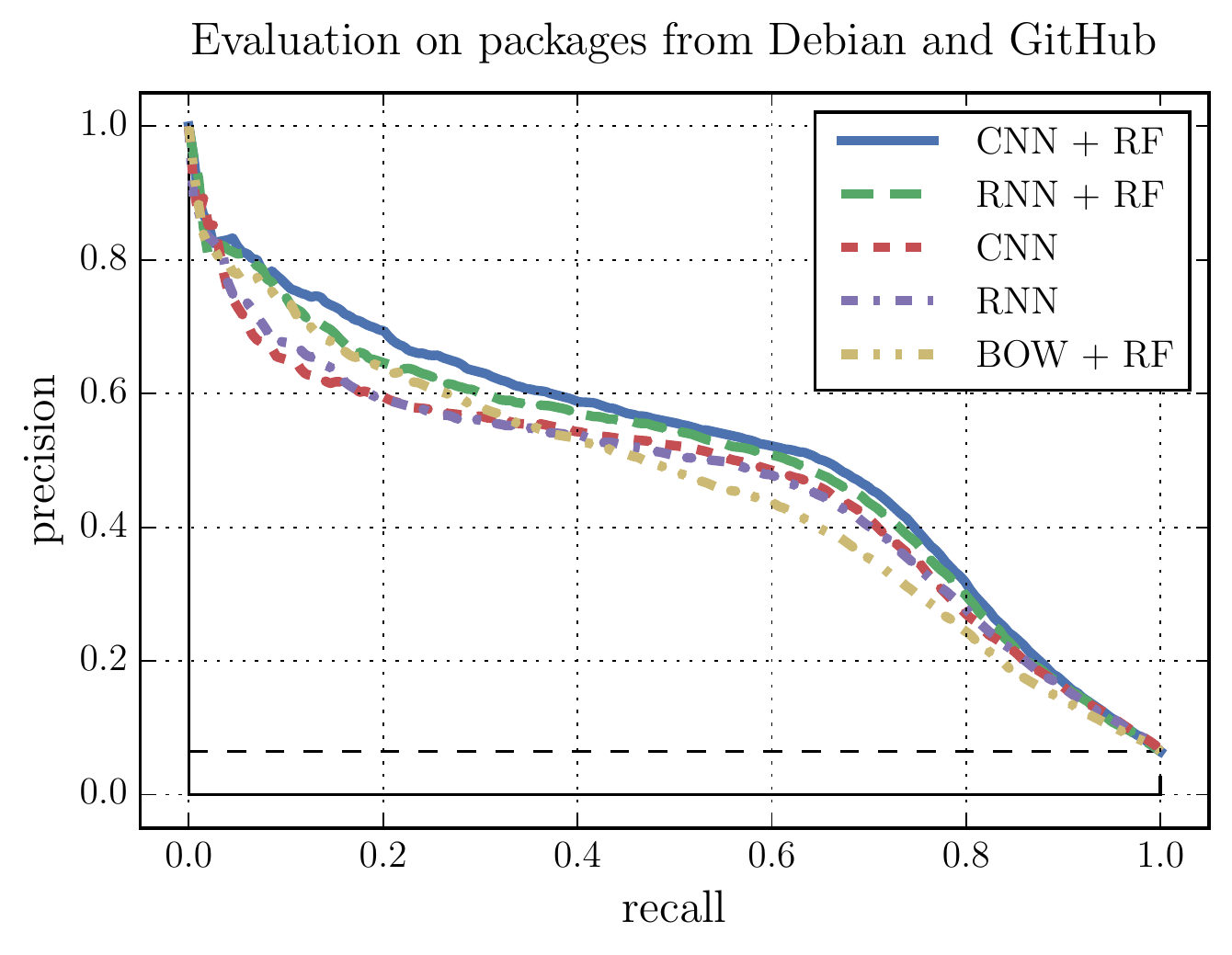}
\caption{Precision versus recall of different ML approaches using our lexer representation on Debian and Github test data. Vulnerable functions make up 6.5\% of the test data.}
\label{fig:test_pr}
\vfill
\end{minipage}
\hfill
\begin{minipage}[t]{0.48\textwidth}
\includegraphics[width=\textwidth]{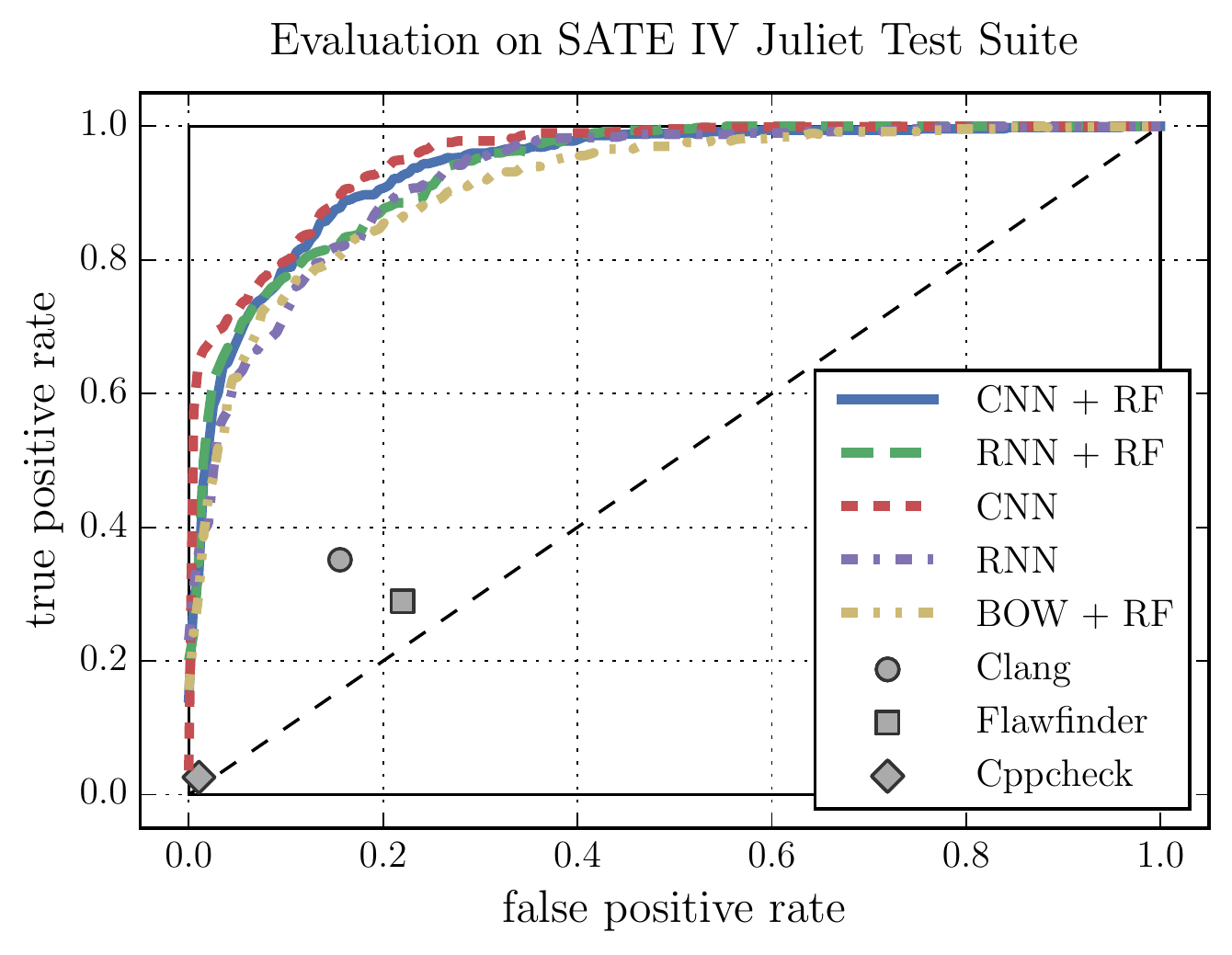}
\caption{SATE IV test data ROC, with true vulnerability labels, compared to the three static analyzers we considered. Vulnerable functions make up 43\% of the test data.}
\label{fig:juliet}
\end{minipage}
\end{figure*}

\begin{figure}
\includegraphics[width=0.48\textwidth]{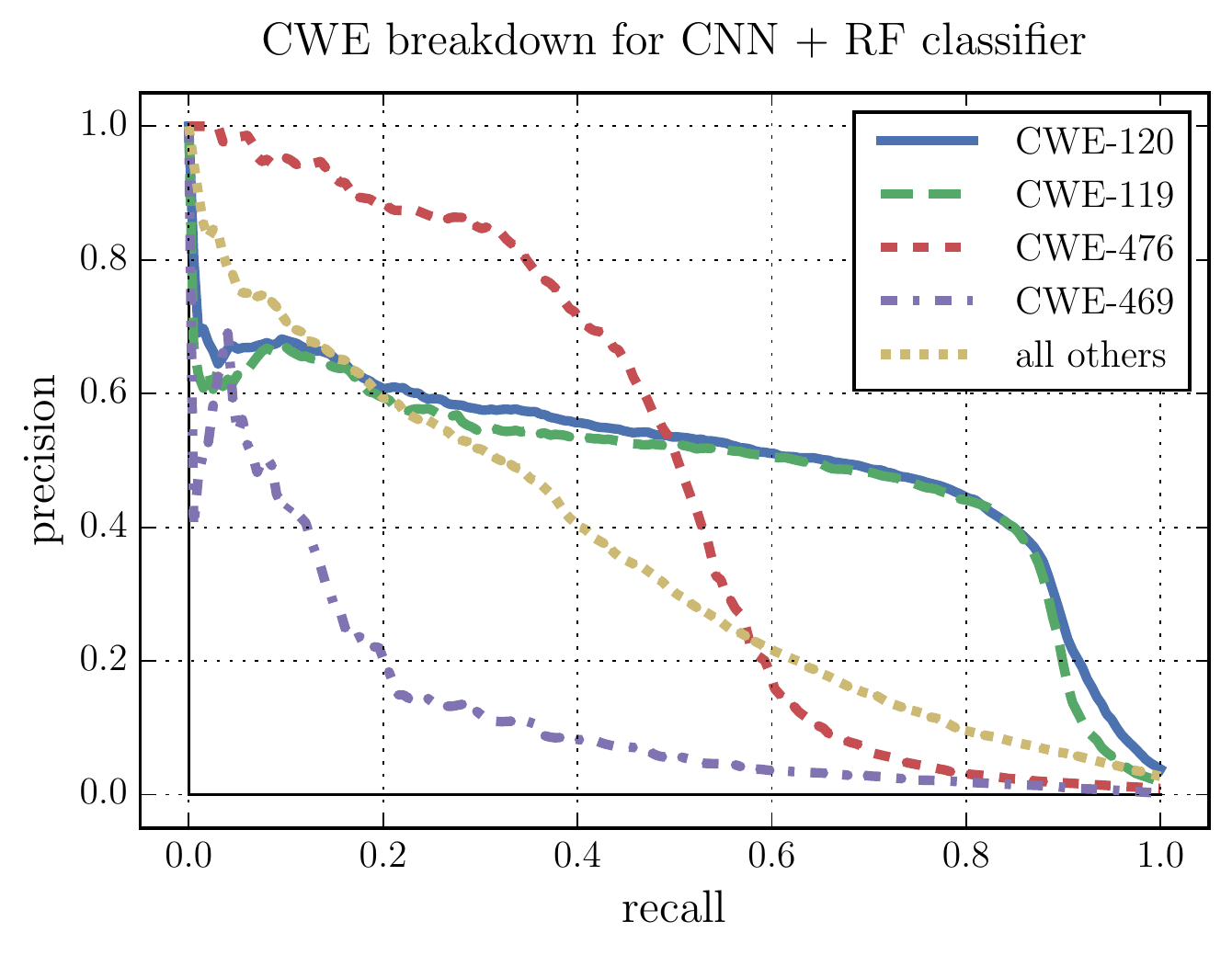}
\caption{Performance of a multi-label CNN + RF classifier on Debian and Github data by vulnerability type (see Table \ref{table:bug_types}.)}
\label{fig:multilabel}
\end{figure}

To provide a strong benchmark, we trained an RF classifier on a ``bag-of-words'' (BOW) representation of function source code, which ignores the order of the tokens. An examination of the bag-of-words feature importances shows that the classifier exploits label correlations with (1) indicators of the source length and complexity and (2) combinations of calls which are commonly misused and lead to vulnerabilities (such as \texttt{memcpy} and \texttt{malloc}.) Improvements over this baseline can be interpreted as being due to more complex and specific vulnerability indication patterns.

Overall, our CNN models performed better than the RNN models as both standalone classifiers and feature generators. In addition, the CNNs were faster to train and required much fewer parameters. On our natural function dataset, the RF classifier trained on neural feature representations performed better than the standalone network for both the CNN and RNN features. Likewise, the RF classifiers trained on neural network representations performed better than the benchmark BOW classifier.

Figure \ref{fig:test_pr} shows the precision-recall performance of the best versions of all of the primary ML approaches on our natural function test dataset. The area under the precision-recall curve (PR AUC) and receiver operating characteristic (ROC AUC) as well as the MCC and $F_1$ score at the validation-optimal thresholds are shown in Table \ref{table:results_test}. Figure \ref{fig:multilabel} shows the performance of our strongest classifier when trained to detect specific vulnerability types from a shared feature representation. Some CWE types are significantly more challenging than others.

\begin{table}
\begin{center}
\begin{tabular}{lcccc}
\toprule
& PR AUC & ROC AUC & MCC & $F_1$ \\
\midrule
BOW + RF & 0.459 & 0.883 & 0.462 & 0.498 \\
RNN & 0.465 & 0.896 & 0.501 & 0.532 \\
CNN & 0.467 & 0.897 & 0.509 & 0.540 \\
RNN + RF & 0.498 & 0.899 & 0.523 & 0.552 \\
CNN + RF & \textbf{0.518} & \textbf{0.904} & \textbf{0.536} & \textbf{0.566} \\
\bottomrule
\end{tabular}
\caption{Results on the Debian and GitHub test data for our ML models, corresponding to Figure \ref{fig:test_pr}.}
\label{table:results_test}
\end{center}
\end{table}

\begin{table}
\begin{center}
\begin{tabular}{lcccc}
\toprule
& PR AUC & ROC AUC & MCC & $F_1$ \\
\midrule
Clang & -- & -- & 0.227 & 0.450 \\
Flawfinder & -- & -- & 0.079 & 0.365 \\
Cppcheck & -- & -- & 0.060 & 0.050 \\
\midrule
BOW + RF & 0.890 & 0.913 & 0.607 & 0.786 \\
RNN & 0.900 & 0.923 & 0.646 & 0.807 \\
CNN & \textbf{0.944} & \textbf{0.954} & \textbf{0.698} & \textbf{0.840} \\
RNN + RF & 0.914 & 0.934 & 0.657 & 0.813 \\
CNN + RF & 0.916 & 0.936 & 0.672 & 0.824 \\
\bottomrule
\end{tabular}
\caption{Results on the SATE IV Juliet Suite test data for our ML models and three static analyzers, as in Figure \ref{fig:juliet}.}
\label{table:results_juliet}
\end{center}
\end{table}

We compare our ML models against our collection of SA tools on SATE IV Juliet Suite dataset, which has true vulnerability labels. Figure \ref{fig:juliet} shows the performance of our models alongside the SA findings on this nearly label-balanced dataset. We find that our models, especially the CNN, perform much better on the SATE IV test data than on the natural functions from Debian and GitHub, likely because SATE IV has many examples for each vulnerability it contains and has fairly consistent style and structure. Among the SA tools, Clang performs the best on the SATE IV data, but still finds very few vulnerabilities compared with all of the ML methods. The full SATE IV results are shown in Table \ref{table:results_juliet}.

\begin{figure}
\includegraphics[width=0.5\textwidth]{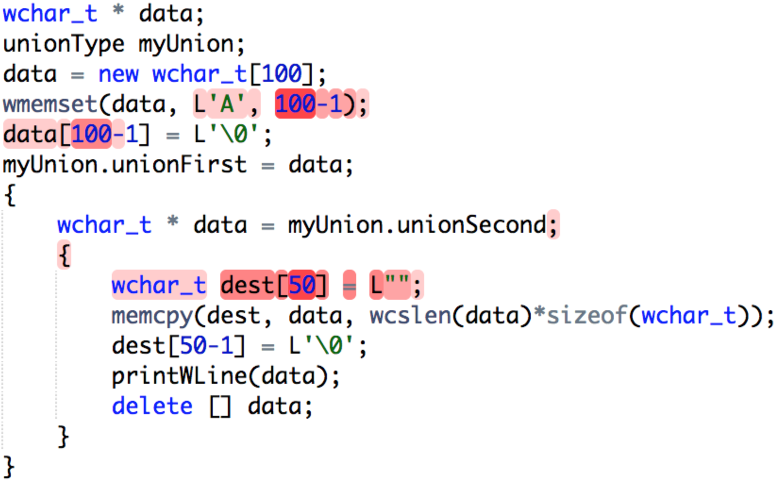}
\caption{Screenshot from our interactive vulnerability detection demo. The convolutional feature activation map \cite{zhou2016} for a detected vulnerability is overlaid in red on the original code.}
\label{fig:viz}
\end{figure}

Our ML methods have some additional advantages over traditional static analysis tools. Our custom lexer and ML models can rapidly digest and score large repositories and source code without requiring that the code be compiled. Additionally, since the ML methods all output probabilities, the thresholds can be tuned to achieve the desired precision and recall. The static analyzers on the other hand return a fixed number of findings, which may be overwhelmingly large for huge codebases or too small for critical applications. While static analyzers are able to better localize the vulnerabilities they find, we can use visualization techniques, such as the feature activation map shown in Figure \ref{fig:viz}, to help understand why our algorithms make their decisions.

%% file: sections/conclusions.tex
We have demonstrated the potential of using ML to detect software vulnerabilities directly from source code. To do this, we built an extensive C/C++ source code dataset mined from Debian and GitHub repositories, labeled with curated vulnerability findings from a suite of static analysis tools, and combined it with the SATE IV dataset. We created a custom C/C++ lexer to create a simple, generic representation of function source code ideal for ML training. We applied a variety of ML techniques inspired by classification problems in the natural language domain, fine-tuned them for our application, and achieved the best overall results using features learned via convolutional neural network and classified with an ensemble tree algorithm.

Future work should focus on improved labels, such as those from dynamic analysis tools or mined from security patches. This would allow scores produced from the ML models to be more complementary with static analysis tools. The ML techniques developed in this work for learning directly on function source code can also be applied to any code classification problem, such as detecting style violations, commit categorization, or algorithm/task classification. As larger and better-labeled datasets are developed, deep learning for source code analysis will become more practical for a wider variety of important problems.

%% file: icmla_2018.bbl
\begin{thebibliography}{10}

\bibitem{cwe}
MITRE, {\em Common Weakness Enumeration}.
\newblock \url{https://cwe.mitre.org/data/index.html}.

\bibitem{BugOccurrence}
T.~D. LaToza, G.~Venolia, and R.~DeLine, ``Maintaining mental models: A study
  of developer work habits,'' in {\em Proc. 28th Int. Conf. Software
  Engineering}, ICSE '06, (New York, NY, USA), pp.~492--501, ACM, 2006.

\bibitem{yadron2014after}
D.~Yadron, ``After heartbleed bug, a race to plug internet hole,'' {\em Wall
  Street Journal}, vol.~9, 2014.

\bibitem{wannacry}
C.~Foxx, ``Cyber-attack: Europol says it was unprecedented in scale.''
  \url{https://www.bbc.com/news/world-europe-39907965}, 2017.

\bibitem{equifax}
C.~Arnold, ``After {E}quifax hack, calls for big changes in credit reporting
  industry.''
  http://www.npr.org/2017/10/18/558570686/after-equifax-hack-calls-for-big-changes-in-credit-reporting-industry,
  2017.

\bibitem{sate4juliet}
NIST, {\em Juliet test suite v1.3}, 2017.
\newblock \url{https://samate.nist.gov/SRD/testsuite.php}.

\bibitem{Xu_Clang}
Z.~Xu, T.~Kremenek, and J.~Zhang, ``A memory model for static analysis of {C}
  programs,'' in {\em Proc. 4th Int. Conf. Leveraging Applications of Formal
  Methods, Verification, and Validation}, pp.~535--548, 2010.

\bibitem{King}
J.~C. King, ``Symbolic execution and program testing,'' {\em Commun. ACM},
  vol.~19, pp.~385--394, July 1976.

\bibitem{Allamanis2017}
M.~Allamanis, E.~T. Barr, P.~T. Devanbu, and C.~A. Sutton, ``A survey of
  machine learning for {B}ig {C}ode and naturalness,'' {\em CoRR},
  vol.~abs/1709.06182, 2017.

\bibitem{Hovsepyan}
A.~Hovsepyan, R.~Scandariato, W.~Joosen, and J.~Walden, ``Software
  vulnerability prediction using text analysis techniques,'' in {\em Proc. 4th
  Int. Workshop Security Measurements and Metrics}, MetriSec '12, pp.~7--10,
  2012.

\bibitem{Pang}
Y.~Pang, X.~Xue, and A.~S. Namin, ``Predicting vulnerable software components
  through n-gram analysis and statistical feature selection,'' in {\em 2015
  IEEE 14th Int. Conf. Machine Learning and Applications (ICMLA)}, 2015.

\bibitem{Mou}
L.~Mou, G.~Li, Z.~Jin, L.~Zhang, and T.~Wang, ``{TBCNN:} {A} tree-based
  convolutional neural network for programming language processing,'' {\em
  CoRR}, 2014.

\bibitem{VulDeePecker}
Z.~Li {\em et~al.}, ``Vul{D}ee{P}ecker: {A} deep learning-based system for
  vulnerability detection,'' {\em CoRR}, vol.~abs/1801.01681, 2018.

\bibitem{harer2018}
J.~Harer {\em et~al.}, ``Learning to repair software vulnerabilities with
  generative adversarial networks,'' {\em arXiv preprint arXiv:1805.07475},
  2018.

\bibitem{debian}
Debian, {\em Debian - the universal operating system}.
\newblock \url{https://www.debian.org/}.

\bibitem{github}
Github, {\em Github}.
\newblock \url{https://github.com/}.

\bibitem{ManyBugs}
C.~{Le Goues} {\em et~al.}, ``The {ManyBugs} and {IntroClass} benchmarks for
  automated repair of {C} programs,'' {\em IEEE Transactions on Software
  Engineering (TSE)}, vol.~41, pp.~1236--1256, December 2015.
\newblock http://dx.doi.org/10.1109/TSE.2015.2454513.

\bibitem{travis}
T.~CI, {\em Travis CI}.
\newblock \url{https://travis-ci.org/}.

\bibitem{Zhou}
Y.~Zhou and A.~Sharma, ``Automated identification of security issues from
  commit messages and bug reports,'' in {\em Proc. 2017 11th Joint Meeting
  Foundations of Software Engineering}, pp.~914--919, 2017.

\bibitem{cppcheck}
Cppcheck, {\em Cppcheck}.
\newblock \url{http://cppcheck.sourceforge.net/}.

\bibitem{flawfinder}
D.~A. Wheeler, {\em Flawfinder}.
\newblock \url{https://www.dwheeler.com/flawfinder/}.

\bibitem{kim2014}
Y.~Kim, ``Convolutional neural networks for sentence classification,'' in {\em
  Proc. 2014 Conf. Empirical Methods in Natural Language Processing (EMNLP)},
  (Doha, Qatar), pp.~1746--1751, Association for Computational Linguistics,
  October 2014.

\bibitem{mikolov2013}
T.~Mikolov, I.~Sutskever, K.~Chen, G.~S. Corrado, and J.~Dean, ``Distributed
  representations of words and phrases and their compositionality,'' in {\em
  Advances in Neural Information Processing Systems}, pp.~3111--3119, 2013.

\bibitem{zhou2016}
B.~Zhou, A.~Khosla, A.~Lapedriza, A.~Oliva, and A.~Torralba, ``Learning deep
  features for discriminative localization,'' in {\em Computer Vision and
  Pattern Recognition (CVPR), 2016 IEEE Conference on}, pp.~2921--2929, IEEE,
  2016.

\end{thebibliography}
